%% file: main.tex
\documentclass{article}
\usepackage{ijcai25}
\usepackage{times}
\usepackage{soul}
\usepackage{url}
\usepackage[hidelinks]{hyperref}
\usepackage[utf8]{inputenc}
\usepackage[small]{caption}
\usepackage{graphicx}
\usepackage{amsmath}
\usepackage{amssymb}
\usepackage{amsthm}
\usepackage{booktabs}
\usepackage{algorithm}
\usepackage{algorithmic}
\usepackage{multirow}
\usepackage[normalem]{ulem}
\useunder{\uline}{\ul}{}
\usepackage[switch]{lineno}
\usepackage{mathtools}
\usepackage{bm}

\usepackage{tikz}

\usepackage[edges]{forest}
\definecolor{hidden-draw}{RGB}{0,0,0}
\definecolor{hidden-pink}{rgb}{0.98, 0.94, 0.75}
\definecolor{level00}{rgb}{0.96, 0.96, 0.86}
\definecolor{level0}{rgb}{0.58, 0.77, 0.98}
\definecolor{level1}{rgb}{0.98, 0.84, 0.69}
\definecolor{level2}{rgb}{0.8, 0.8, 1.0}
\definecolor{level3}{rgb}{1.0, 0.71, 0.76}
\definecolor{level4}{rgb}{0.49, 0.99, 0.0}

\definecolor{lawngreen}{rgb}{0.49, 0.99, 0.0}
\definecolor{pink}{rgb}{1, 0, 0.5}
\definecolor{airforce}{rgb}{0.36, 0.54, 0.66}

\hypersetup{
    linkbordercolor=airforce,
    urlbordercolor=pink,
    citebordercolor=green,
}
\title{Towards Cross-Modality Modeling for Time Series Analytics: A Survey in the LLM Era}

\author{
Chenxi Liu$^1$\and
Shaowen Zhou$^1$\and
Qianxiong Xu$^1$\thanks{Corresponding author}\and
Hao Miao$^2$\and
Cheng Long$^{1*}$\and
Ziyue Li$^{3*}$\and\\
Rui Zhao$^4$
\\
\affiliations
$^1$S-Lab, Nanyang Technological University, Singapore\\
$^2$Aalborg University, Denmark\\
$^3$University of Cologne, Germany\\
$^4$SenseTime Research, China\\
\emails
    \{chenxi.liu, qianxiong.xu, c.long\}@ntu.edu.sg, s200061@e.ntu.edu.sg\\haom@cs.aau.dk, zlibn@wiso.uni-koeln.de, zhaorui@sensetime.com
}
\begin{document}

\maketitle

\begin{abstract}
The proliferation of edge devices has generated an unprecedented volume of time series data across different domains, motivating various well-customized methods. Recently, Large Language Models (LLMs) have emerged as a new paradigm for time series analytics by leveraging the shared sequential nature of textual data and time series. However, a fundamental cross-modality gap between time series and LLMs exists, as LLMs are pre-trained on textual corpora and are not inherently optimized for time series. Many recent proposals are designed to address this issue. In this survey, we provide an up-to-date overview of LLMs-based cross-modality modeling for time series analytics. We first introduce a taxonomy that classifies existing approaches into four groups based on the type of textual data employed for time series modeling. We then summarize key cross-modality strategies, e.g., alignment and fusion, and discuss their applications across a range of downstream tasks. Furthermore, we conduct experiments on multimodal datasets from different application domains to investigate effective combinations of textual data and cross-modality strategies for enhancing time series analytics. Finally, we suggest several promising directions for future research. This survey is designed for a range of professionals, researchers, and practitioners interested in LLM-based time series modeling.
\end{abstract}

\section{Introduction}

% ========================================
% Paragraph 1
% Time Series Analysis
% ========================================
% {\hao 
With the proliferation of edge devices and the development of mobile sensing techniques, a large amount of time series data has been generated, enabling a variety of real-world applications~\cite{liu2025csur,pettersson2023time,liu2024spatial,DBLP:journals/www/CaiWCLX24,chenxi2021study}. Time series data typically take the format of sequential observations with varying features~\cite{liu2024mvcar,DBLP:conf/dsaa/AlnegheimishNBV24,liu2024icde,DBLP:journals/www/LiuXWCCC22}. 
Considerable research efforts have been made to design time series modeling and analytics methods, which enables different downstream tasks, such as 
%Time series analysis encompasses processing these numeric data for multiple tasks,
%including 
time series forecasting~\cite{21-TNSE-Liu,DBLP:conf/hpcc/ChenWL20,JIN2022315}, imputation~\cite{chen2024llm,DBLP:journals/tits/XiaoX0HLZ022}, classification~\cite{liu2024taming,liu2021understanding}, and anomaly detection~\cite{xu2024pefad}.

% ========================================
% Paragraph 2
% LLMs for Time Series Analysis
% The key challenge - cross-modality gap
% Shortcomings of exsiting survey
%========================================
% {\hao 
Recently, large language model (LLM)-based methods~\cite{touvron2023llama,radford2019language} have emerged as a new paradigm for time series modeling. These methods are inspired by time series and natural text exhibit similar formats (i.e., sequence)~\cite{DBLP:conf/dexa/YangSLLMLZ24}, and assume that the generic knowledge learned by LLMs can be easily transferred to time series~\cite{xue2023promptcast}.
Although existing surveys have introduced broad overviews of LLM-based time series methods~\cite{jin2024position,DBLP:conf/ijcai/0001C0S24,jiang2024empowering}, they overlook the critical challenge posed by the \textit{cross-modality gap}~\cite{liu2024taming} between time series and textual data. To be specific, LLMs are pre-trained on textual corpora and are not inherently designed for the time series, there is a pressing need to develop cross-modality modeling strategies that effectively integrate textual knowledge into time series analytics.

\begin{figure}[t] 
\centering 
\includegraphics[width=0.5\textwidth]{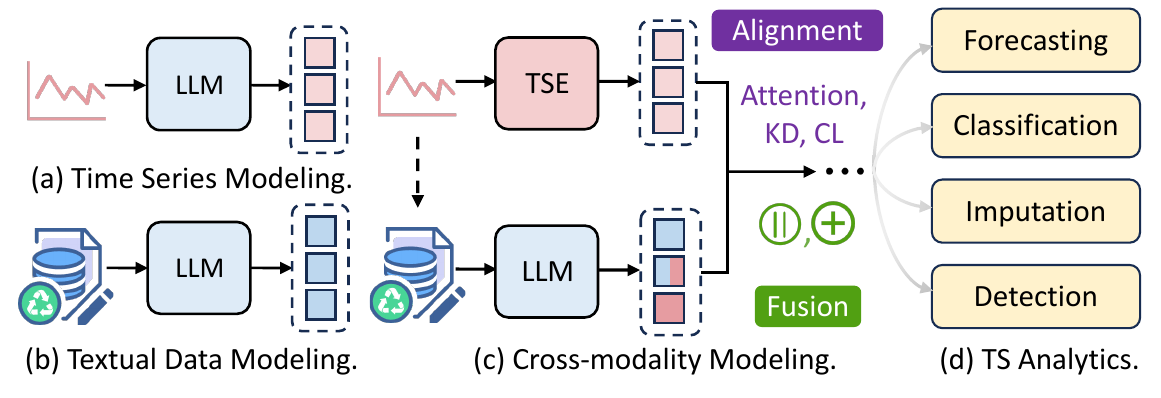} 
\caption{Cross-Modality Modeling for Time Series Analytics.}
\label{fig:intro} 
\end{figure}

% ========================================
% Paragraph 3
% Our contribution
% ========================================
% {\hao 
% To address this issue, 
This survey makes a unique contribution to the existing literature by addressing the cross-modality gap between time series and textual data, thereby enhancing LLM-based time series analytics. % with the integration of LLMs. 
%As illustrated in Figure~\ref{fig:intro}, we focus on time series modeling that incorporates textual data, emphasizing cross-modality modeling methods. 
Figure~\ref{fig:intro} shows a general framework for LLM-based time series modeling. In this paper, we divide textual data into four types: \textit{numerical prompt, statistical prompt, contextual prompt, and word token embedding}. To contend with the cross-modality modeling, we summarize two overarching strategies according to recent studies\cite{jin2023time,liu2024timecma}, i.e., alignment and fusion, to integrate time series with different textual data.
For alignment, we identify four key methods: \textit{unidirectional retrieval, bidirectional retrieval, contrastive learning, and knowledge distillation}. In addition, the fusion strategy primarily relies on \textit{concatenation} and/or \textit{addition} to integrate textual information into time series embeddings. Furthermore, this survey spans diverse application domains, including healthcare, electricity, economics, weather, and traffic, showcasing the broad applicability of the proposed taxonomy. Finally, we conduct experimental evaluations on multi-domain multimodal datasets to assess the effective combinations of textual data and cross-modality strategies for effective time series forecasting, providing practical insights for future research.

The major contributions are summarized as follows.
\begin{itemize}
    \item We present a comprehensive catalog of literature on LLM-based cross-modality modeling for time series analytics, highlighting recent representative methods.
    \item We propose a taxonomy that classifies related studies into four groups based on the type of textual data. Additionally, we explore cross-modality modeling strategies, including alignment and fusion, and discuss their applications across various tasks and domains.
    \item We perform experimental evaluations on multi-domain multimodal datasets to explore effective combinations of additional textual data and strategies that facilitate time series analytics.
\end{itemize}

\section{Formulation}
\subsection{Definitions}
\paragraph{Time Series.} We define a time series as an ordered sequence, denoted by $\mathbf{X} = \left\{\mathbf{x}_{1}, \ldots, \mathbf{x}_{S}\right\} \in \mathbb{R}^{S \times N}$, where $S$ represents the sequence length, and $N$ is the number of variables. Each observation $\mathbf{x}_i$ is an $N$-dimensional vector at time step $i$. The scalar $v_i$ refers to the numerical value of a specific variable in the time series at time step $i$.

\paragraph{Textual Data.}
The textual data $\mathbf{T} = \{\mathbf{P}, \mathbf{W}\}$ in time series modeling can be categorized into four types: numerical prompt~\cite{DBLP:conf/nips/GruverFQW23}, statistical prompt~\cite{liu2024timemmd}, contextual prompt~\cite{liu2024unitime}, and word token embedding~\cite{pan2024s}. Some studies~\cite{liu2024timecma,jin2023time} utilize a combination of prompts, denoted as $\mathbf{P}=\{\mathbf{P}_{\text{N}}, \mathbf{P}_{\text{S}}, \mathbf{P}_{\text{C}}\}$, while others directly adopt word token embeddings $\mathbf{W}$~\cite{pan2024s,liu2024time} extracted from LLMs. In this survey, we unify the definitions of these textual data types as follows:

\begin{itemize}
    \item \textbf{Numerical Prompt} transforms the numerical data of $\mathbf{X}$ into a textual format, denoted as $\mathbf{P}_{\text{N}}$. Each prompt consists of $M$ words, primarily representing the numerical values of the time series.

    \item \textbf{Statistical Prompt} encodes statistical features of the time series, such as mean, maximum, minimum, median, top-$k$, or trend values. These statistics are typically expressed in textual format and denoted as $\mathbf{P}_{S}$.

    \item \textbf{Contextual Prompt} provides auxiliary descriptions, including dataset metadata, media news, or event-related information. We denote contextual instructions as $\mathbf{P}_{\text{C}}$.

    \item \textbf{Word Token Embedding} refers to the pre-trained weights within LLMs. Instead of using textual prompts, the textual representations can be directly captured from the word token embeddings, denoted as $\mathbf{W}$.
\end{itemize}

\subsection{Cross-Modality Modeling for Time Series Analytics}
Given time series $\mathbf{X}$ and textual data $\mathbf{T}$, cross-modality modeling aims to learn a function that integrates $\mathbf{X}$ with $\mathbf{T}$ to generate the target output $\mathbf{Y}$ for downstream tasks, such as long-term forecasting, short-term forecasting, classification, imputation, and anomaly detection.
Formally, the objective is to learn a mapping function:
\begin{equation}
    f: (\mathbf{X}, \mathbf{T}) \rightarrow \mathbf{Y},
\end{equation}
where $f(\cdot)$ is the method that aligns and fuses both modalities to enhance time series modeling.
% $\mathbf{T} = \{\mathbf{P}, \mathbf{W}\}$ represents the textual modality, which can be either prompts $\mathbf{P}$ or word token embeddings $\mathbf{W}$. The function $f(\cdot)$ is the method that aligns and fuses both modalities to enhance time series modeling.

% \subsection{Cross-Modality Time Series Modeling}
% Given time series data $\mathbf{X}$ and textual data $\mathbf{P}$, $\mathbf{W}$. The cross-modality time series analysis aims to learn a function using $\mathbf{X}$ with $\mathbf{P}$ or $\mathbf{W}$ to generate the target $\mathbf{Y}$ for downstream tasks, such as long-term forecasting, short-term forecasting, imputation, classification, anomaly detection, and reasoning.

% such as long-term forecasting~\cite{liu2024timecma}, short-term forecasting~\cite{huang2024leret}, imputation~\cite{chen2024llm}, classification~\cite{liu2024taming}, anomaly detection~\cite{DBLP:conf/dsaa/AlnegheimishNBV24}, and reasoning~\cite{tao2024hierarchical}.

\input{img/cma_taxonomy}

\begin{figure}[h] 
\centering 
\includegraphics[width=0.45\textwidth]{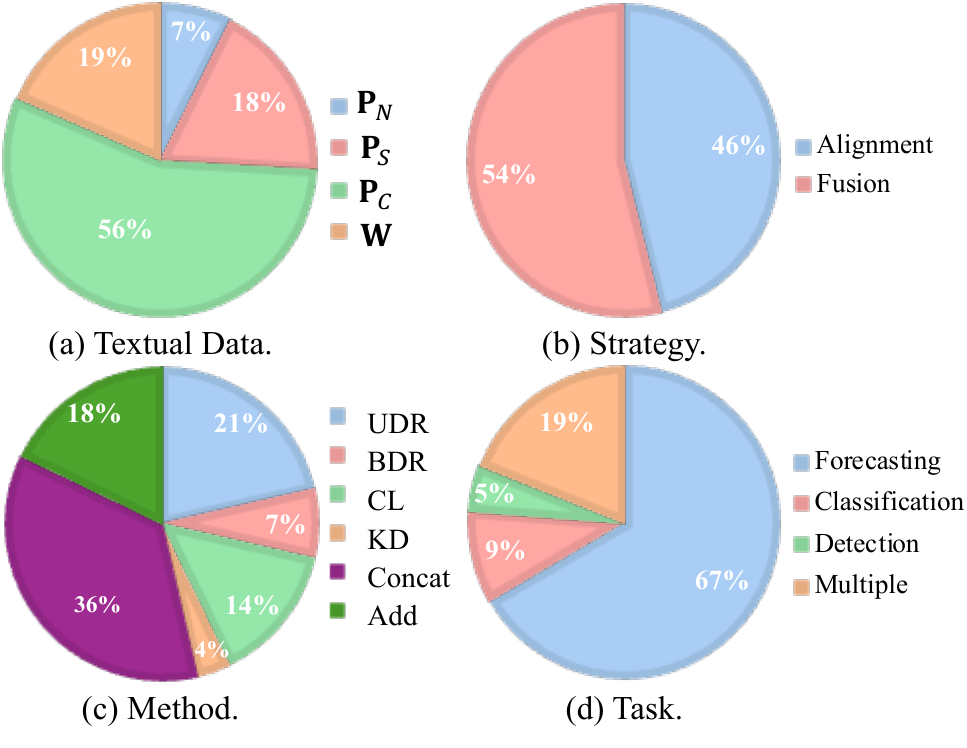} 
\caption{Distribution of taxonomies in cross-modality time series modeling. (a) Textual data types: numerical prompts $\mathbf P_N$, statistical prompts $\mathbf P_S$, contextual prompts $\mathbf P_C$, and word token embeddings $\mathbf{W}$. (b) Strategy: Alignment vs. Fusion. (c) Method categories: unidirectional retrieval (UDR), bidirectional retrieval (BDR), contrastive learning (CL), knowledge distillation (KD), concatenation (Concat), and addition (Add). (d) Task types: forecasting, classification, anomaly detection, and multiple tasks.}
\label{fig:Distribution} 
\end{figure}

% \subsection{Taxonomy Explanation}

\section{Cross-Modality Alignment}
This section presents cross-modality alignment, which aims to learn the association of time series and textual data. We highlight three widely adopted alignment methods: retrieval, contrastive learning, and knowledge distillation.

\subsection{Retrieval}
Retrieval is the method of leveraging data from one modality to access relevant information in another. Based on the retrieval direction, we categorize it into two types: unidirectional retrieval, where information flows from one modality to another, and bidirectional retrieval, where both modalities can retrieve information from each other.

\subsubsection{Unidirectional Retrieval}
% In unidirectional retrieval, information from one modality is leveraged to enhance the modeling of another modality. 
This method has been applied to forecasting tasks across general domains. For example, TimeCMA~\cite{liu2024timecma} introduces hybrid prompts that integrate numerical, statistical, and contextual information to improve time series forecasting. These hybrid prompts are processed by an LLM to generate prompt embeddings, which are then aligned with the original time series through unidirectional similarity-based retrieval. This retrieval process leverages time series embeddings to extract disentangled and robust time series representations from the LLM-empowered prompt embeddings.

Similarly, Time-LLM~\cite{jin2023time}, Time-FFM~\cite{liu2024time}, $S^2$IP-LLM~\cite{pan2024s}, and CALF~\cite{liu2024taming} employ unidirectional retrieval by aligning time series embeddings with word token embeddings in pre-trained LLMs, using the former as queries. In contrast, TEMPO~\cite{cao2023tempo} takes an inverse approach, utilizing prompt embeddings as queries to retrieve the top-$K$ corresponding values from the patched time series input.

Overall, the unidirectional retrieval method typically involves using the time series embedding $\mathbf{E}_X$ to retrieve relevant information from the LLM-enhanced textual embedding $\mathbf{E}_T$, implemented via cross-attention:
\begin{equation}
\mathbf{E}_A = \mathit{CrossAttention}(\mathbf Q, \mathbf K, \mathbf V),
\end{equation}
\noindent where $\mathbf{E}_A$ is the aligned time series embedding.
\begin{equation}
\mathbf Q = \mathbf{E}_X \mathbf W_Q, \quad K = \mathbf{E}_T \mathbf W_K, \quad \mathbf V = \mathbf{E}_T \mathbf W_V.
\end{equation}

Conversely, if the prompt embeddings act as the query:
\begin{equation}
\mathbf Q = \mathbf{E}_T \mathbf W_Q, \quad \mathbf K = \mathbf{E}_X \mathbf W_K, \quad V = \mathbf{E}_X \mathbf W_V,
\end{equation}

\noindent Here, $\mathbf W_Q, \mathbf W_K, \mathbf W_V$ are learnable projection matrices that transform the features into query, key, and value spaces.

\subsubsection{Bidirectional Retrieval}
This method extends unidirectional retrieval by allowing both time series and textual embeddings to retrieve information from each other. 
This method ensures deeper cross-modality interactions and enhances downstream tasks, including forecasting and classification, across multiple domains.

For instance, LeRet~\cite{huang2024leret} introduces a bidirectional retrieval method for time series forecasting. Instead of relying solely on time series embeddings as queries, LeRet allows textual embeddings to retrieve relevant time series features, creating a dynamic exchange between the two modalities. This bidirectional retrieval strategy improves forecasting accuracy by leveraging the strengths of both data sources.

In the healthcare domain, DualTime~\cite{zhang2024dualtime} employs bidirectional retrieval for forecasting and classification, integrating textual and time series embeddings to enhance predictive modeling in clinical applications. Formally, bidirectional retrieval can be expressed as an extension of unidirectional retrieval, where either modality can act as the query. For example, textual knowledge is mapped to the time series feature space in the first stage:
\begin{equation}
\mathbf{E}_T' = \mathit{CrossAttention}(\mathbf{E}_T, \mathbf{E}_X, \mathbf{E}_X).
\end{equation}

Second stage aims to integrate this aligned textual knowledge with time series features:
\begin{equation}
\mathbf{E}_A = \mathit{CrossAttention}(\mathbf{E}_X, \mathbf{E}_T', \mathbf{E}_T'),
\end{equation}

\noindent where $\mathbf{E}_T'$ denotes the aligned textual embeddings.

\subsection{Contrastive Learning}
Contrastive learning aims to establish a shared representation by maximizing the agreement between corresponding time series and textual embeddings while minimizing the similarity between non-corresponding pairs~\cite{DBLP:conf/iclr/OzyurtF023}. For example, Chen et al.~\cite{chen2024llm} propose a contrastive module to align time series and textual prompts by maximizing the mutual information between small model’s time series representation and LLM’s textual representation. 
% HiTime~\cite{tao2024hierarchical} presents a dual-view contrastive alignment module that bridges the gap between modalities via facilitating semantic space alignment between time series and contextual prompt.

Similarly, METS~\cite{li2024frozen} utilizes the auto-generated clinical reports to guide electrocardiogram (ECG) self-supervised pre-training.
% The authors use a trainable encoder and a frozen LLM to embed paired ECG and automatically reports separately. 
The contrastive stragegy aims to maximize the similarity between paired and report while minimize the similarity between ECG and other reports. 
TEST~\cite{DBLP:conf/iclr/Sun0LH24} builds an encoder to embed TS via instance-wise, feature-wise, and text-prototype-aligned contrast, where the TS embedding space is aligned to LLM’s embedding layer space.

Formally, contrastive learning for cross-modality alignment can be defined as follows. Given a time series embeddings $\mathbf{E}_X$ and textual embeddings $\mathbf{E}_T$, the contrastive loss function is formulated as:
\begin{equation}
\mathcal{L}_{\text{contrast}} = - \log \frac{\exp \left( \text{sim}(\mathbf{E}_X, \mathbf{E}_T)/\tau \right)}{\sum_{\mathbf{\bar{E}}_T \in \mathcal{N}} \exp \left( \text{sim}(\mathbf{E}_X, \mathbf{\bar{E}}_T)/\tau \right)},
\end{equation}
where $\text{sim}(\cdot, \cdot)$ denotes the similarity function (e.g., cosine similarity), $\tau$ is a temperature hyperparameter, and $\mathcal{N}$ represents a set of negative samples (i.e., unrelated textual embeddings). The objective is to maximize the similarity between aligned pairs $(\mathbf{E}_X, \mathbf{E}_T)$ while minimizing the similarity between mismatched pairs $(\mathbf{E}_X, \mathbf{\bar{E}}_T)$.

\subsection{Knowledge Distillation}
The LLM-based knowledge distillation (KD) achieves a small student model from an LLM, enabling efficient inference solely on the distilled student model.
Recent works have been proposed to address the cross-modal misalignment problem with knowledge distillation~\cite{liu2025crossST}, which can generally be categorized into black-box distillation~\cite{liu2024large} and white-box distillation\cite{liu2025timekd} based on the accessibility of the teacher model’s internal information during the distillation process.

CALF~\cite{liu2024taming} is a black-box KD method that aligns LLMs for time series forecasting via cross-modal fine-tuning. 
To adapt the word token embeddings to time series data, they align the outputs of each intermediate layer $l$ in
the time series-based LLM with those of the textual LLM, also aligns the output consistency between these two modalities to maintain a coherent semantic representation:
\begin{equation}
\mathcal{L}_{\text{feature}}=\sum_{i=1}^L \gamma^{(L-i)} \operatorname{sim}\left(\mathbf F_{X}^l, \mathbf F_{T}^l\right),
\mathcal{L}_{\text{output }}=\operatorname{sim}\left(\mathbf E_{X}, \mathbf E_{T}\right),
\end{equation}
where $\mathbf F_{X}^l$ and $\mathbf F_{T}^l$ are the outputs of the $l$-th Transformer block in time series-based and textual LLMs, respectively. $L$ is the total number of layers in the LLM. $\gamma$ is the hyperparameter that controls the loss scale from different layers.

In contrast, TimeKD~\cite{liu2025timekd} is a white-box KD method benefits the design of privileged correlation distillation, the student model explicitly aligns its internal attention maps with those of the teacher model to mimic their behavior.
% CALF also aligns the output consistency between these two modalities to maintain a coherent semantic representation:
% \begin{equation}
% \mathcal{L}_{\text{output }}=\operatorname{sim}\left(\mathbf E_{X}, \mathbf E_{T}\right).
% \end{equation}

\section{Cross-Modality Fusion}
This section introduces cross-modality fusion, which refers to the process of combining textual and time series data into a union representation. Fusion strategy allows models to leverage complementary information from different modalities~\cite{DBLP:journals/csur/ZhaoZG24}, enhancing their ability to capture richer contextual dependencies. We summarize two common fusion methods: concatenation and addition of embeddings. Unlike alignment strategies, fusion-based methods often introduce data entanglement issues~\cite{liu2024timecma}, which may lead to suboptimal performance compared to alignment-based methods.

\subsection{Addition}
Addition-based fusion integrates textual embeddings with time series representations by summing their feature vectors. This method allows models to incorporate textual information without significantly increasing the dimensionality of the feature space, making it a computationally efficient alternative to concatenation. Unlike concatenation, addition-based fusion maintains a compact representation, ensuring that the model does not introduce unnecessary complexity while still leveraging multimodal information.

Several studies have adopted addition-based fusion for time series analysis. Time-MMD~\cite{liu2024timemmd}, GPT4MTS~\cite{jia2024gpt4mts}, AutoTimes~\cite{liu2024autotimes}, and T3~\cite{han2024event} add the textual embedding with time series embedding for time series analysis, respectively. Formally, the addition can be expressed as follows:
\begin{equation}
    \mathbf{E}_F = \mathbf{E}_X~+~\mathbf{E}_T,
\end{equation}
where $\mathbf{E}_F$ is the fused embeddings and $+$ denotes addition.

\subsection{Concatenation}
Concatenation-based fusion directly merges textual embeddings with time series features to create a joint representation. This method enables models to incorporate textual information alongside time series data, allowing for a more comprehensive feature space. While concatenation provides a straightforward way to multimodal integration, it can increase the dimensionality of the feature space, leading to greater computational complexity. Moreover, the lack of explicit alignment mechanisms between modalities may introduce noise, reducing the effectiveness of downstream tasks.

% single Concatenation-based fusion strategy
Some studies directly concatenate time series and textual embeddings. For instance, UniTime~\cite{liu2024unitime} concatenates the contextual prompt embedding with time series embedding to retained a LLM-based unified model for cross-domain time series forecasting.
SIGLLM~\cite{DBLP:conf/dsaa/AlnegheimishNBV24} concatenates the contextual prompt embedding with time series embedding for zero-shot anomaly detection task.
TEMPO~\cite{cao2023tempo} concatenates the word token embedding with different time series feature, such as trend, seasonal, and residual, for time series forecasting.
Time-FFM~\cite{liu2024time} concatenates word token embedding with time series embedding for time series forecasting. The concatenation can be formulated as follows:
\begin{equation}
    \mathbf{E}_F = \mathbf{E}_X~||~\mathbf{E}_T.
\end{equation}
where $||$ denotes concatenation.

% multiple strategies
Other works utilize multiple strategies to integrate data embeddings.
Beyond retrieval-based alignment, Time-LLM~\cite{jin2023time} further enhances the adaptability of LLMs for time series forecasting by concatenating the textual prompt embedding as a prefix to the general time series embedding. $S^2$IP-LLM~\cite{pan2024s} concatenates time series embedding and retrieved embedding to avoid the data entanglement issue.
FCSA~\cite{hu2025context} concatenates the time series embeddings and textual prompt embeddings to extract fine-grained features for further alignment.
After addition-based fusion, T3~\cite{han2024event} concatenates the contextual prompt and traffic data embeddings to maximize the utilization of the training data for the traffic forecasting task.

\input{section/5_experiment}

\section{Future Directions}
% Figure~\ref{fig:taxonomy} presents a taxonomy of cross-modality models for time series analytics that integrates textual data via LLMs. 
% Based on the preceding review, we believe there is still much space for further innovation in this field.

\paragraph{Multi-Modality Modeling.} Expanding beyond the integration of time series and textual data, future research could delve into additional modalities such as images~\cite{huang2024multi}, video~\cite{wang2024condition}, and audio~\cite{huang2021audio}. In this context, it is essential to explore the capability of LLMs for enhancing multi-modality representation. Recent advancements in multi-modal LLMs exemplify the potential of such integrations. For instance, Meta's LlaMa 3.2 processes both images and textual data, enabling applications ranging from augmented reality to document summarization. 
% Similarly, Google's Gemini 2.0 is designed to handle text, images, audio, and video, facilitating a wide array of applications across different domains. 
These developments underscore the importance of investigating how LLMs can be leveraged to create effective multi-modality modeling.

\paragraph{Improving Effectiveness.} While LLM-based cross-modality methods have demonstrated strong capabilities, they do not always surpass smaller, task-specific models~\cite{DBLP:conf/nips/WangWDQZLQWL24}. In some cases, employing an LLM with an excessive number of parameters can lead to overfitting, particularly on specialized tasks across several domains. Future research could focus on techniques such as dynamic model selection, meta-learning, and continual adaptation that can help improve model effectiveness by allowing models to adjust to changing data distribution.

\paragraph{Efficient Optimization.} Despite their success, existing studies still meet the challenge of high computational costs, particularly when processing long sequences, more tokens, or handling multivariate data. This is due to the high dimensionality of multivariate time series (i.e., multiple variables over timestamps) and the multi-head attention mechanism within LLMs
Recent advancements have explored strategies to mitigate this challenges, such as last token storage~\cite{liu2024timecma}, knowledge distillation~\cite{DBLP:conf/iclr/Gu0WH24}. Future research could focus on developing lightweight architectures, efficient attention mechanisms, and adaptive computation frameworks to optimize efficiency and scalability.

\paragraph{Transparency of LLMs.} LLMs have demonstrated remarkable performance in textual–time series analytics~\cite{wang2024news}, yet they often operate as “black-box” systems, raising concerns about their reasoning processes and overall transparency. Much of the current research primarily applies or fine-tunes LLMs without an explicit focus on exposing their internal reasoning processes. This lack of interpretability can hinder trust, particularly in high-stakes applications such as healthcare and finance. Moreover, LLMs are prone to generating hallucinations, seemingly plausible but incorrect outputs, which further complicates their deployment in real-world scenarios. Future research on textual–time series analysis could prioritize enhancing the transparency of LLMs, ensuring that these models operate more reliably during subsequent alignment or fusion processes.

\section{Conclusion}
This paper aims to highlight the importance of cross-modality modeling for time series analytics in the LLM era. We propose a novel taxonomy from a textual data-centric perspective, categorizing existing studies by key data types, namely numerical prompts, statistical prompts, contextual prompts, and word token embeddings. Our main premise is through cross-modality alignment and fusion, textual data can significantly enhance time series analytics tasks across diverse domains. To validate this viewpoint, we perform multi-domain multimodal experiments to systematically evaluate the effectiveness of various alignment and fusion strategies in key time series tasks. Finally, we explore open challenges and promising directions for future research.

\section{Acknowledgments}
This study is supported under the RIE2020 Industry Alignment Fund – Industry Collaboration Projects (IAF-ICP) Funding Initiative, as well as cash and in-kind contributions from the industry partner(s).
% \section*{Ethical Statement}

% There are no ethical issues.

% \section*{Acknowledgments}

%% The file named.bst is a bibliography-style file for BibTeX 0.99c
\bibliographystyle{named}
\bibliography{mmtsa}

\end{document}

%% file: img/cma_taxonomy.tex
\tikzstyle{my-box}=[
    rectangle,
    draw=hidden-draw,
    rounded corners,
    text opacity=1,
    minimum height=1.5em,
    minimum width=5em,
    inner sep=2pt,
    align=center,
    fill opacity=.5,
    line width=0.8pt,
]
\tikzstyle{leaf}=[my-box, minimum height=1.5em,
    fill=hidden-pink!80, text=black, align=left, font=\normalsize,
    inner xsep=2pt,
    inner ysep=4pt,
    line width=0.8pt,
]

\begin{figure*}[!t]
    \centering
    % \begin{tikzpicture}
    %     % 图例块
    %     % \node[draw, fill=level0!80, fill=level0, rounded corners, minimum width=3cm, minimum height=1cm, font=\normalsize, align=center] at (0, 0) {\textbf{Foundation Models}};
    %     \node[draw, fill=level1!80, rounded corners, minimum width=1.8cm, minimum height=0.5cm, font=\normalsize, align=center] at (4.2, 0) 
    %     {\textbf{Textual Data}};
    %     \node[draw, fill=level2!60, rounded corners, minimum width=1.5cm, minimum height=0.5cm, font=\normalsize, align=center] at (6.5, 0) {\textbf{Strategy}};
    %     \node[draw, fill=level3!60, rounded corners, minimum width=2.5cm, minimum height=0.5cm, font=\normalsize, align=center] at (8.8, 0) {\textbf{Method}};
    %     \node[draw, fill=level4!20, rounded corners, minimum width=1.5cm, minimum height=0.5cm, font=\normalsize, align=center] at (11.3, 0) {\textbf{Task}};
    %     \node[draw, fill=hidden-pink!60, rounded corners, minimum width=2cm, minimum height=0.5cm, font=\normalsize, align=center] at (13.4, 0) {\textbf{Domain}};
    % \end{tikzpicture}

    % \vspace{0.4cm} % 调整图例与分类树之间的间距
    \resizebox{\textwidth}{!}
    {
        \begin{forest}
            forked edges,
            for tree={
                fill=level0!80,
                grow=east,
                reversed=true,
                anchor=base west,
                parent anchor=east,
                child anchor=west,
                base=left,
                font=\large,
                rectangle,
                draw=hidden-draw,
                rounded corners,
                align=left,
                minimum width=4em,
                edge+={darkgray, line width=1pt},
                s sep=4pt,
                % s sep=3pt, % 调整垂直间距
                % l sep=12pt, % 调整水平间距
                inner xsep=2pt,
                inner ysep=3pt,
                line width=0.8pt,
                ver/.style={rotate=90, child anchor=north, parent anchor=south, anchor=center},
            },
            where level=1{text width=9em,font=\normalsize,fill=level1!80,}{},
            where level=2{text width=10em,font=\normalsize,fill=level2!80,}{},
            where level=3{text width=8em,font=\normalsize,fill=level3!60,}{},
            where level=4{text width=10em,font=\normalsize,fill=level4!20,}{},
            where level=5{text width=5em,font=\normalsize,}{},
            % [\textbf{Time Series}\\\textbf{Analytics}\\\textbf{with Text}, align=center, fill=level00
            [\textbf{TS Analytics with}\\\textbf{Textual Data}, align=center, fill=level00
            % [TS as Text, text width=4.5em, align=center
                % [
                % \textbf{General}: LLMTIME~\cite{gruver2023llmtime}.
                % ]
            % ]
            % [~~w/ Text, text width=4.5em, align=center
                [~Numerical\\~Prompt, text width=5em, align=center
                    [Alignment, text width=4.5em
                        [Unidirectional Retrieval, text width=10em
                            [Forecasting, text width=5.5em
                                [\textbf{General}: TimeCMA~\cite{liu2024timecma}, leaf, text width=17em
                                ]
                            ]
                        ]
                        [Contrastive Learning, text width=10em
                            [Multiple, text width=5.5em
                                [\textbf{General}: LLM-TSI~\cite{chen2024llm}, leaf, text width=17em
                                ]
                            ]
                        ]         
                        [Knowledge Distillation, text width=10em
                            [Forecasting, text width=5.5em
                                [\textbf{General}: TimeKD~\cite{liu2025timekd}, leaf, text width=17em
                                ]
                            ]
                        ]
                    ]
                ]       
                [~Statistical\\~Prompt, text width=5em, align=center
                    [Alignment, text width=4.5em
                        [Unidirectional Retrieval, text width=10em
                            [Forecasting, text width=5.5em
                                [\textbf{General}: TimeCMA~\cite{liu2024timecma}, leaf, text width=17em
                                ]
                            ]
                            ]
                        [Contrastive Learning, text width=10em
                            [Multiple, text width=5.5em
                                [\textbf{General}: LLM-TSI~\cite{chen2024llm}, leaf, text width=17em
                                ]
                            ]
                        ]         
                    ]
                    [Fusion, text width=4.5em
                            [Concatenation, text width=10em
                                [Forecasting, text width=5.5em
                                    [\textbf{General:} Time-LLM~\cite{jin2023time}, leaf, text width=17em]]
                                    [Multiple, text width=5.5em
                                    [\textbf{General:} FSCA~\cite{hu2025context}, leaf, text width=17em
                                    ]]    
                            ]
                            [Addition, text width=10em
                                [Forecasting, text width=5.5em
                                [
                                \textbf{General:} Time-MMD~\cite{liu2024timemmd}, leaf, text width=17em
                                ]
                            ]    
                            ]
                        ]
                    ]
                    [~Contextual\\Prompt, text width=5em, align=center
                        [Alignment, text width=4.5em
                            [Unidirectional Retrieval, text width=10em
                                [Forecasting, text width=5.5em
                                    [\textbf{General}: TimeCMA~\cite{liu2024timecma}, leaf, text width=17em]
                                ]
                            ]
                            [Bidirectional Retrieval, text width=10em
                                [Forecasting, text width=5.5em
                                    [\textbf{General:} LeRet~\cite{huang2024leret}, leaf, text width=17em ]
                                ]
                                [Multiple, text width=5.5em
                                    [\textbf{Health:} DualTime~\cite{zhang2024dualtime}, leaf, text width=17em
                                    ]
                                ]
                            ]
                                % [Contrastive Learning, text width=10em
                        [Contrastive Learning, text width=10em
                            [Classification, text width=5.5em
                                    [\textbf{General:} HiTime~\cite{tao2024hierarchical}, leaf, text width=17em
                                    ]
                                    [\textbf{Health:} METS~\cite{li2024frozen}, leaf, text width=17em ]
                            ]
                            [Multiple, text width=5.5em
                                [\textbf{General}: LLM-TSI~\cite{chen2024llm}, leaf, text width=17em
                                ]
                            ]
                        ]
                        [Knowledge Distillation, text width=10em
                            [Forecasting, text width=5.5em
                                [\textbf{General}: TimeKD~\cite{liu2025timekd}, leaf, text width=17em
                                ]
                            ]
                        ]
                            % ]
                    ]
                    [Fusion, text width=4.5em
                            [Concatenation, text width=10em
                                [Forecasting, text width=5.5em
                                    [\textbf{General:} Time-LLM~\cite{jin2023time}{,}\\UniTime~\cite{liu2024unitime}, leaf, text width=17em
                                    ]
                                    [\textbf{Traffic:} T3~\cite{han2024event}, leaf, text width=17em ]
                                ]
                                [Detection, text width=5.5em
                                [\textbf{General:} SIGLLM~\cite{DBLP:conf/dsaa/AlnegheimishNBV24}, leaf, text width=20em
                                    ]
                                ]
                                [Multiple, text width=5.5em
                                    [\textbf{General:} FSCA~\cite{hu2025context}, leaf, text width=17em
                                    ]
                                ]
                            ]
                            [Addition, text width=10em
                                [Forecasting, text width=5.5em
                                    [\textbf{General:} AutoTimes~\cite{liu2024autotimes}{,}\\Time-MMD~\cite{liu2024timemmd}, leaf, text width=17em ]
                                    [\textbf{Event:} GPT4MTS~\cite{jia2024gpt4mts}, leaf, text width=17em
                                    ]
                                    [\textbf{Traffic:} T3~\cite{han2024event}, leaf, text width=17em ]
                                ]
                            ]
                    ]
                    ]
                    [Word Token\\Embedding, text width=5em, align=center
                        [Alignment, text width=4.5em
                            [Unidirectional Retrieval, text width=10em
                                [Forecasting, text width=5.5em
                                    [\textbf{General:} Time-LLM~\cite{jin2023time}{,}\\ TEMPO~\cite{cao2023tempo}
                                    {,}\\Time-FFM~\cite{liu2024time}{,}\\$S^2$IP-LLM~\cite{pan2024s}
                                    {,}\\CALF~\cite{liu2024taming}
                                    , leaf, text width=17em
                                    ]
                                ]
                            ]
                            [Contrastive Learning, text width=10em
                                [Forecasting, text width=5.5em
                                    [\textbf{General:} TEST~\cite{DBLP:conf/iclr/Sun0LH24}, leaf, text width=17em ]
                                ]
                            ]
                            [Knowledge Distillation, text width=10em
                                [Multiple, text width=5.5em
                                    [\textbf{General:} CALF~\cite{liu2024taming}, leaf, text width=17em ]
                                ]
                            ]]
                        [Fusion, text width=4.5em
                            [Concatenation, text width=10em
                            [Forecasting, text width=5.5em
                        [\textbf{General:} TEMPO~\cite{cao2023tempo}{,}\\Time-FFM~\cite{liu2024time}{,}\\$S^2$IP-LLM~\cite{pan2024s}, leaf, text width=17em ]]]
                        ]
                ]
            % ]
            % [~w/o Text, text width=4.5em
                % [
                % \textbf{General}: PeFAD~\cite{xu2024pefad}{,} AnomalyLLM~\cite{liu2024large}{,}  OFA~\cite{DBLP:conf/nips/ZhouNW0023}, leaf, text width=41.5em
                % ]
                % [
                % \textbf{Traffic}: STD-PLM~\cite{huang2024std}{,} ST-LLM~\cite{liu2024spatial}{,} TrafficBERT~\cite{DBLP:journals/eswa/JinWLKKK21}, leaf, text width=41.5em
                % ]
                % % \textbf{General}: PeFAD~\cite{xu2024pefad}{,} AnomalyLLM~\cite{liu2024large}{,}  OFA~\cite{DBLP:conf/nips/ZhouNW0023}{,} GATGPT~\cite{chen2023gatgpt}, leaf, text width=52.6em
                % % ]
                % % [
                % % \textbf{Traffic}: STD-PLM~\cite{huang2024std}{,} ST-LLM~\cite{liu2024spatial}{,} TrafficBERT~\cite{DBLP:journals/eswa/JinWLKKK21}{,} TPLLM~\cite{ren2024tpllm}, leaf, text width=52.6em
                % % ]
            % ]
            ]
        \end{forest}
}    
\caption{Taxonomy of cross-modality modeling for time series (TS) analytics incorporating textual data, including numerical prompt, statistical prompt, contextual prompt, and word token embedding. The textual data is processed by LLMs.}
\label{fig:taxonomy}
\end{figure*}
% , categorized according to methodologies (i.e., unconditioned vs. conditioned), tasks (e.g., predictive versus generative), and applications.
% Single-Modality
%  [xxx, leaf, text width=22em ]

% others： 展开
% 

%% file: section/5_experiment.tex
\section{Experiments}
We perform extensive experimental evaluations on multi-domain, multimodal datasets. We employ four types of textual data in Figure~\ref{fig:Distribution}(a) as well as cross-modality alignment and fusion strategies in Figure~\ref{fig:Distribution}(b). Specifically, we select the three most common methods from the literature: unidirectional retrieval-based alignment (21\%), concatenation-based fusion (36\%), and addition-based fusion (18\%) in Figure~\ref{fig:Distribution}(c). We focus on the time series forecasting task, accounting for 67\% of reported tasks across multiple domains in Figure~\ref{fig:Distribution}(d). We also implement a single-modality model without textual inputs. Our code and datasets are available\footnote{\url{https://github.com/ChenxiLiu-HNU/CM2TS}}.
% Through empirical analysis, we aim to determine which types of textual data are most beneficial and which alignment or fusion methods are most effective for time series forecasting.

\subsection{Dataset Description}
We utilize datasets from five domains~\cite{liu2024timemmd}, spanning agriculture, climate, economy, energy, and health. Each dataset consists of a univariate time series paired with relevant textual data.
 The textual data includess expert reports and news summaries, each annotated with timestamps corresponding to the periods they describe.

\paragraph{Time Series Data} is summarized in Table~\ref{tab:dataset}, covering five domains: agriculture, climate, economy, energy, and health. The datasets are recorded at varying temporal resolutions, including weekly and monthly frequencies, with records spanning from the 1980s to 2024. 
Each dataset consists of univariate or multivariate time series, with the number of dimensions ranging from 1 to 11.

\begin{table}[t]
\centering
\caption{Overview of datasets.}
\begin{tabular}{lcccl}
\toprule
Domain      & Dim & Frequency & Samples & Timespan \\ 
\midrule
Agriculture & 1         & Monthly   & 496     & 1980 - 2024 \\ 
Climate     & 5         & Monthly   & 496     & 2000 - 2024 \\ 
Economy     & 3         & Monthly   & 423     & 1987 - 2024 \\ 
Energy      & 9         & Weekly    & 1479    & 1993 - 2024 \\ 
Health      & 11        & Weekly    & 1389    & 2002 - 2024 \\ 
\bottomrule
\end{tabular}
\label{tab:dataset}
\end{table}

\begin{figure}[t] 
\centering 
\includegraphics[width=0.4\textwidth]{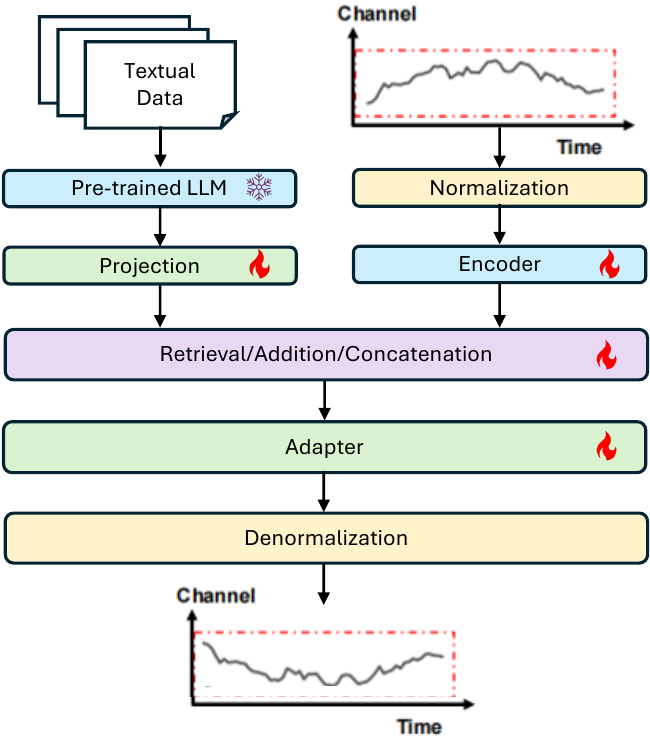} 
\caption{Overall Framework.} 
\label{fig:model} 
\end{figure}

\paragraph{Textual Data} consists of expert reports and news summaries. Expert reports are categorized as statistical prompts, as they provide insights into averages and trends within specific timeframes. News summaries can serve as contextual prompts, offering cues about future trends. We follow TimeCMA~\cite{liu2024timecma} to wrap time series values into numerical prompts. The word token embeddings are obtained from the pre-trained GPT-2~\cite{pan2024s}.

% offering qualitative information that indicate potential future trends.
% To integrate textual data with time series, we manually align texts with timestamps. Within a given lookback window, we concatenate texts corresponding to relevant timestamps to ensure proper temporal alignment.
% data process
%% drop TS without associated texts

\subsection{Implementation Details}

\begin{table*}[t]
\centering
\resizebox{\textwidth}{!}{
\begin{tabular}{l|c|cc|cc|cc|cc|cc}
\toprule
\multirow{2}{*}{Data} & \multirow{2}{*}{Method} 
  & \multicolumn{2}{c|}{Agriculture} 
  & \multicolumn{2}{c|}{Climate} 
  & \multicolumn{2}{c|}{Energy} 
  & \multicolumn{2}{c|}{Economy} 
  & \multicolumn{2}{c}{Health} \\

 & & MSE & MAE & MSE & MAE & MSE & MAE & MSE & MAE & MSE & MAE \\ 
\midrule
\multicolumn{12}{c}{Single Modality} \\ 
\midrule
Time Series & -
  & 2.68 & 1.35 
  & 0.371 & 0.473 
  & 0.183 & 0.315 
  & 0.0244 & 0.128 
  & 1.12 & 0.774 \\
\midrule
\multicolumn{12}{c}{Cross Modality} \\ 
\midrule
\multirow{3}{*}{\begin{tabular}[c]{@{}l@{}}Time Series~\&\\ Numerical Prompt\end{tabular}} 
  & Retrieval 
    & 2.83 & 1.38 
    & \textbf{0.287} & \textbf{0.425} 
    & 0.186 & 0.315 
    & 0.0247 & 0.129 
    & \textbf{0.96} & \textbf{0.665} \\
  & Addition 
    & 2.77 & 1.34 
    & 0.297 & 0.434 
    & \textbf{0.180} & \textbf{0.309} 
    & 0.0252 & 0.132 
    & 1.05 & 0.754 \\
  & Concatenation 
    & 2.87 & 1.41 
    & {\ul 0.296} & {\ul 0.425}
    & 0.224 & 0.350 
    & 0.0267 & 0.132 
    & 1.16 & 0.797 \\
\midrule

\multirow{3}{*}{\begin{tabular}[c]{@{}l@{}}Time Series~\&\\ Statistical Prompt\end{tabular}} 
  & Retrieval 
    & {\ul2.67} & {\ul1.33} 
    & 0.386 & 0.483 
    & 0.196 & 0.333 
    & \textbf{0.0232} & \textbf{0.125} 
    & {\ul 0.97} & {\ul 0.667} \\
  & Addition 
    & \textbf{2.64} & \textbf{1.29} 
    & 0.380 & 0.478 
    & 0.183 & 0.313 
    & {\ul 0.0244} & {\ul 0.126} 
    & 1.08 & 0.771 \\
  & Concatenation 
    & 2.75 & 1.38 
    & 0.386 & 0.488 
    & 0.190 & 0.325 
    & 0.0254 & 0.130 
    & 1.16 & 0.802 \\
\midrule

\multirow{3}{*}{\begin{tabular}[c]{@{}l@{}}Time Series~\&\\ Contextual Prompt\end{tabular}} 
  & Retrieval 
    & 2.88 & 1.35 
    & 0.389 & 0.493 
    & 0.185 & 0.317 
    & 0.0261 & 0.131 
    & 1.11 & 0.661 \\
  & Addition 
    & 2.85 & 1.32 
    & 0.364 & 0.467 
    & {\ul 0.182} & {\ul 0.313} 
    & 0.0263 & 0.132 
    & 1.17 & 0.713 \\
  & Concatenation 
    & 2.93 & 1.41 
    & 0.387 & 0.485 
    & 0.193 & 0.327 
    & 0.0272 & 0.134 
    & 1.28 & 0.776 \\
\midrule

\multirow{3}{*}{\begin{tabular}[c]{@{}l@{}}Time Series~\&\\ Word Token Embedding\end{tabular}} 
  & Retrieval & 2.90 & 1.37 
   & 0.388 & 0.490 
   & 0.181 & 0.312 
   & 0.0271 & 0.135 
   & 1.18 & 0.721 \\
 & Addition 
  & 2.95 & 1.45 
  & 0.393 & 0.479 
  & 0.187 & 0.319 
  & 0.0265 & 0.133 
  & 1.22 & 0.768 \\
 & Concatenation 
  & 2.92 & 1.42 & 
  0.389 & 0.484 & 
  0.188 & 0.320 & 
  0.0274 & 0.138 & 
  1.29 & 0.779 \\
\bottomrule
\end{tabular}}
\caption{Time series forecasting performance across multiple domains using diverse textual data and cross-modality modeling methods.}
% The best result in each domain is shown in \textbf{bold}, and the second best is \ul{underlined}.}
\label{tab:results}
\end{table*}

Figure \ref{fig:model} presents an overview of our cross-modality modeling framework for time series analysis, comprising three key components: a pre-trained LLM, an alignment or fusion layer, and an adapter, detailed below.

\paragraph{Pre-trained LLM.} This module includes a tokenizer and a pre-trained GPT-2 model with frozen weights, efficiently embeds textual data.

\paragraph{Encoder.} It can be a Transformer-based encoder than embeds time series data and captures temporal dynamics~\cite{DBLP:conf/iclr/LiuHZWWML24}, with reversible instance normalization applied to normalize time series values.

\paragraph{Alignment or Fusion.}
% To reduce cross-modality gap, we consider both alignment and fusion methods.
The alignment strategy employs a unidirectional retrieval method.
The fusion strategy includes concatenation and addition methods.
% We first utilize average pooling to obtain a one-text-token embedding. Then, we concatenate or add it to the time series embedding.

%% task adapter
\paragraph{Task Adapter.} The adapter is designed by the downstream task. In this section, we conduct the forecasting task.
We use a linear projection layer to predict the future time series. We also de-normalize the values to restore the actual prediction.

\subsection{Experiment Settings}
We configure the lookback window size to 36 for weekly data and 8 for monthly data, while setting the prediction horizon to 24 time steps across all datasets.
We evaluate models' performance using Mean Squared Error (MSE) and Mean Absolute Error (MAE). The model is optimized using the Adam optimizer, with Cosine Annealing as the learning rate scheduler.
Training is conducted on NVIDIA A100 GPUs, with 20 epochs for weekly data and 50 epochs for monthly data.

\subsection{Results and Discussion}

% Table \ref{tab:results} shows the results of our experiments. We conclude that textual data enhance time series forecasting. Texts that include numerical values such as numerical prompt or statistical prompt work better. The retrieval-based alignment methods and addition-based fusion methods tend to perform better than concatenation-based fusion methods.

Table~\ref{tab:results} presents the results of cross-modality time series forecasting across five domains. The findings highlight the impact of incorporating textual data and provide insights into the effectiveness of different alignment and fusion methods.

\paragraph{Textual Data Enhances Foresting.}
Overall, textual data significantly enhance time series forecasting performance. For example, in the climate domain, the retrieval-based numerical prompt reduces MSE by 22.6\% compared to the time series-only baseline.
Among the different types of textual information, numerical prompts and statistical prompts lead to the most notable improvements. These text types contain numerical values that directly correlate with time series patterns, providing structured signals for forecasting. In contrast, contextual prompts and word token embeddings show relatively weaker performance, as they contain less structured information, which may not align as closely with numerical trends.

\paragraph{Numerical Prompts Perform Better.}
Numerical prompts consistently deliver the best performance across most domains. This is particularly evident in climate, energy, and health forecasting. For example, in the climate and health domains, numerical prompts reduce MSE by 21.15\% and 17.95\%, respectively, compared to contextual prompts. Statistical prompts show strong results in economy forecasting. Word token embeddings perform the worst, indicating that general semantic representations may not effectively capture time series-relevant information.

\paragraph{Alignment Outperforms Fusion.}
We evaluate three methods for each textual data type: retrieval-based alignment, addition-based fusion, and concatenation-based fusion. Retrieval-based alignment consistently performs well, particularly for numerical and statistical prompts. In the the economy domain, when using numerical prompts as textual data, retrieval reduces MSE by 16.96\% compared to concatenation. Addition-based fusion often outperforms concatenation with the three textual prompts but underperforms it when using word token embeddings.

\paragraph{Domain-Specific Observations.}  
The impact of textual data varies across different domains. Economy and health forecasting benefit considerably from numerical and statistical prompts, where structured reports align well with economic indicators and public health trends.
Climate and agriculture forecasting show more moderate improvements. The reason is that these domains may rely on more external factors that are not always well captured by textual descriptions.
Energy forecasting achieves the better results with addition-based fusion of numerical prompts.

% In ssummary, our findings emphasize the importance of structured numerical information for cross-modality time series forecasting. Retrieval-based alignment and addition-based fusion consistently yield better results。